\documentclass[letterpaper]{article} 
\usepackage{aaai2026}  
\usepackage{times}  
\usepackage{helvet}  
\usepackage{courier}  
\usepackage[hyphens]{url}  
\usepackage{graphicx} 
\usepackage{natbib}  
\usepackage{caption} 
\usepackage{algorithm}
\usepackage{algorithmic}
\usepackage{xcolor}

\usepackage{newfloat}

\usepackage{amsmath}
\usepackage{amssymb}
\usepackage{graphicx}

\usepackage{listings}
\usepackage{threeparttable}

\usepackage{multirow}
\usepackage{booktabs}
\usepackage{makecell} 

\DeclareCaptionStyle{ruled}{labelfont=normalfont,labelsep=colon,strut=off} 
\floatstyle{ruled}
\newfloat{listing}{tb}{lst}{}
\floatname{listing}{Listing}
\title{Beyond ReAct: A Planner-Centric Framework for Complex \\ Tool-Augmented LLM Reasoning }

\author{
    Xiaolong Wei\textsuperscript{\rm 1}\thanks{Equal contribution.},
    Yuehu Dong\textsuperscript{\rm 2}\footnotemark[1],
    Xingliang Wang\textsuperscript{\rm 3},
    Xingyu Zhang\textsuperscript{\rm 4},
    Zhejun Zhao\textsuperscript{\rm 2}\thanks{Corresponding author.},
    Dongdong Shen\textsuperscript{\rm 2},
    Long Xia\textsuperscript{\rm 2},
    Dawei Yin\textsuperscript{\rm 2}
}
\affiliations{
    \textsuperscript{\rm 1}Beihang University\\
    \textsuperscript{\rm 2}Baidu Inc.\\
    \textsuperscript{\rm 3}Beijing University of Posts and Telecommunications\\
    \textsuperscript{\rm 4}Beijing Jiaotong University\\
    
    xiaolongwei@buaa.edu.cn, 
    wangxingliang@bupt.edu.cn, 
    zhangxingyu@bjtu.edu.cn, \\
    \{dongyuehu, zhaozhejun, shendongdong\}@baidu.com, 
    long.phil.xia@gmail.com, 
    yindawei@acm.org
}

\usepackage{bibentry}

\begin{document}

\maketitle

\begin{abstract}
Existing tool-augmented large language models (LLMs) encounter significant challenges when processing complex queries. Current frameworks such as ReAct are prone to local optimization traps due to their reliance on incremental decision-making processes. To address these limitations, we propose a novel Planner-centric Plan-Execute paradigm that fundamentally resolves local optimization bottlenecks through architectural innovation. Central to our approach is a novel Planner model that performs global Directed Acyclic Graph (DAG) planning for complex queries, enabling optimized execution beyond conventional tool coordination. We also introduce ComplexTool-Plan, a large-scale benchmark dataset featuring complex queries that demand sophisticated multi-tool composition and coordination capabilities. Additionally, we develop a two-stage training methodology that integrates Supervised Fine-Tuning (SFT) with Group Relative Policy Optimization (GRPO), systematically enhancing the Planner's tool selection accuracy and global planning awareness through structured DAG-based planning. When integrated with a capable executor, our framework achieves state-of-the-art performance on the StableToolBench benchmark for complex user queries, demonstrating superior end-to-end execution capabilities and robust handling of intricate multi-tool workflows. Our code and data are publicly available at \url{https://github.com/weixiaolong94-hub/Beyond-React}.
\end{abstract}

\section{Introduction} 
Large Language Models (LLMs) have demonstrated remarkable prowess in language processing, yet their inherent knowledge is static and they lack direct interaction with the external world. A transformative solution is tool augmentation, which empowers LLMs to act as autonomous agents by calling external APIs and tools \citep{schick2023toolformer, patil2024gorilla}. This paradigm shifts LLMs from being mere text generators to active problem-solvers capable of tackling complex, real-world tasks, a trend exemplified by systems like HuggingGPT \citep{shen2023hugginggpt} and the rise of multi-agent frameworks \citep{hong2023metagpt}. The evaluation of these agentic capabilities has itself become a major research thrust, with new benchmarks like AgentBench \citep{liu2023agentbench} setting the stage for more rigorous assessment.

\begin{figure}[h]
\centering
\includegraphics[width=\columnwidth]{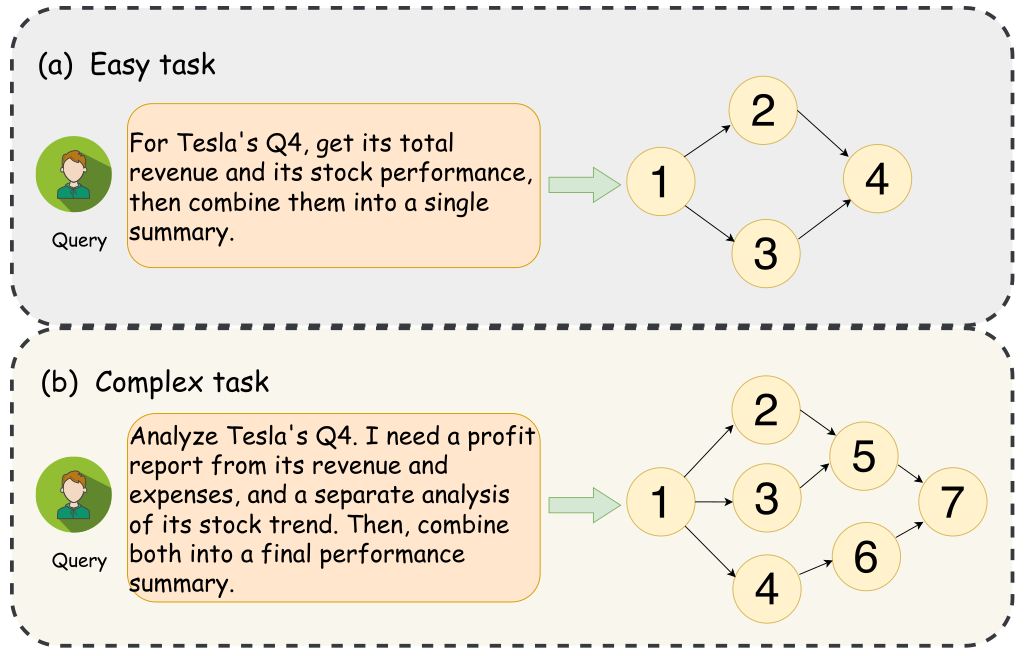} 
\caption{An example of a simple versus a complex task. A simple query results in a basic, parallel DAG, while a complex query involving nested logic is translated into a more elaborate, multi-level DAG.}
\label{fig:example_of_complex}
\end{figure}
\label{sec:intro}

\begin{figure*}[t]
\centering
\includegraphics[width=1\textwidth]{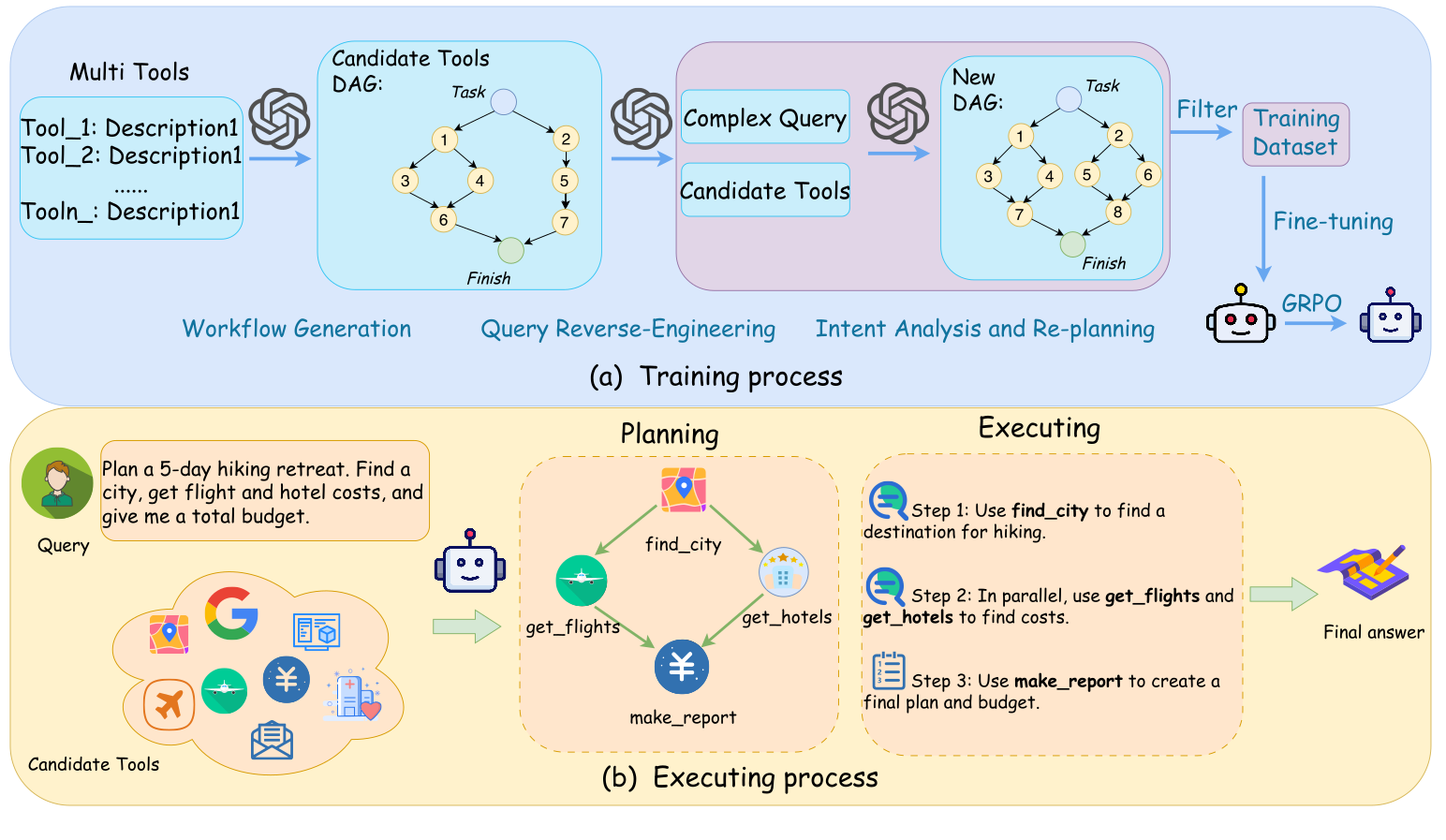} 
\caption{The figure illustrates our proposed framework. (a) The Training Process shows our automated pipeline for creating a training dataset and then training the Planner model via fine-tuning and reinforcement learning (GRPO). (b) The Executing Process demonstrates how the trained Planner takes a user query, generates a parallelizable execution plan as a Directed Acyclic Graph (DAG), and orchestrates the tools to produce the final answer.}
\label{fig:global_framework}
\end{figure*}

The dominant approach for orchestrating tool usage relies on reactive, step-by-step reasoning frameworks like ReAct \citep{yao2023react}, often augmented by self-reflection techniques \citep{shinn2023reflexion}. However, this paradigm suffers from inherent local optimization traps due to its incremental decision-making process. While potentially effective for simple queries, its reactive nature often falters on complex tasks. As illustrated in Fig.~\ref{fig:example_of_complex}, complex tasks involve more intricate dependencies than simple ones. To mitigate this, advanced methods employ tree search \citep{yao2023tree} or explicit planning prompts \citep{wang2023plan} to explore multiple paths. While these methods improve upon purely linear reasoning, they fail to architecturally resolve the fundamental bottleneck of local optimization. Critically, they remain focused on finding an optimal sequential path, overlooking opportunities for parallelism inherent in complex workflows. Moreover, the computational overhead of constructing and traversing search trees, requiring numerous LLM calls, can be prohibitive.

To address this fundamental challenge, we adopt a framework that decouples planning from execution \citep{gao2023pal, chen2022program}. Our approach employs a dedicated Planner model, trained to translate complex user queries into an execution plan structured as a DAG. In this DAG, nodes represent selected tools, and directed edges capture dependency relationships between them. By optimizing the Planner to improve the accuracy of both node selection (tool choices) and edge prediction (dependency relationships) in the DAG, we achieve more effective plans for complex queries, improving workflow execution compared to sequential models.

Realizing such a global planner presents two significant hurdles. First, there is a lack of large-scale, structured data required to train a model to generate complex plans, a challenge addressed by recent large-scale tool-use benchmarks like ToolBench \citep{qin2023toolllm} and API-Bank \citep{li2023api}. Second, evaluating the quality of a plan is non-trivial. To overcome these, we introduce \textbf{ComplexTool-Plan}, a new, large-scale benchmark with graded difficulty levels. It serves as both the training ground and the evaluation standard for sophisticated planning agents. We then devise a two-stage hybrid training strategy for our Planner, combining Supervised Fine-Tuning (SFT) with subsequent refinement via Reinforcement Learning (RL) \citep{ouyang2022training}, a strategy that aligns with recent findings on using self-play and RL to elicit stronger reasoning \citep{chen2024self,wei2025igniting}.

Our extensive experiments validate the efficacy of our decoupled approach. On ComplexTool-Plan, our Planner substantially outperforms leading baselines. Crucially, when our Planner is paired with a powerful executor model like GPT-4o, the integrated system establishes a new state-of-the-art (SOTA) on the widely recognized end-to-end benchmark, StableToolBench \citep{guo2024stabletoolbench}, underscoring the practical superiority of our framework.

In summary, our key contributions are:
\begin{itemize}
    \item We enhance task success in tool-augmented LLMs by proposing a planner-centric plan-execute paradigm framework and developing a DAG-optimized Planner, which significantly improves the accuracy of tool selection (nodes) and dependency modeling (edges).
    \item We construct ComplexTool-Plan, a large-scale, difficulty-graded benchmark designed to foster and evaluate complex agentic planning capabilities.
    \item Through extensive experiments, we demonstrate that our framework achieves state-of-the-art performance, offering a more scalable, efficient, and robust solution for complex multi-tool orchestration.
\end{itemize}

\section{Related Works}
We survey the rapidly evolving landscape of tool-augmented LLMs\citep{li2025rankexpert,li2025rankelectra} along three primary axes: the evolution of tool-calling paradigms, RL-based policy optimization, and the development of evaluation benchmarks.

\subsection{Tool Calling Paradigms}
The paradigms for integrating LLMs with external tools have evolved from the interleaved "Reasoning-Acting" steps of ReAct \citep{yao2023react} to more reliable, natively supported structured API calls. A significant branch of research leverages code as a universal tool interface, with works like Program-of-Thought (PoT) \citep{chen2022program} and PAL \citep{gao2023pal} offloading complex computations to code interpreters. This trend has produced highly specialized agents like SWE-agent \citep{yang2024swe}, which can autonomously resolve software engineering tasks. For instance, advanced models are increasingly required for challenging real-world applications like multi-modal fake news detection, which involves reasoning over diverse data from various sources and domains \citep{tong2025dapt, lu2025dammfnd}.

Recent work is also moving beyond general-purpose models towards those specifically fine-tuned for tool use\citep{zhao2025turatoolaugmentedunifiedretrieval}. For instance, Toolformer \citep{schick2023toolformer} demonstrates that a language model can teach itself to use tools by learning to insert beneficial API calls into text, thus significantly improving its ability to handle complex, tool-dependent instructions in a zero-shot manner. Concurrently, the interaction paradigm is expanding to visual environments, with powerful GUI agents like CogAgent \citep{hong2024cogagent} demonstrating the ability to understand and interact with graphical user interfaces. Beyond reactive or single-agent frameworks, Li et al. \citep{li2025towards} introduce an \textit{AI Search Paradigm} that formalizes reasoning as coordinated planning among multiple LLM-based agents, while Chen et al. \citep{chen2025multi} propose a proactive multi-agent orchestration framework for complex information-seeking tasks. Our work contributes to this evolution by focusing on the upfront, holistic planning of tool interactions, a critical component for efficient orchestration.

\subsection{RL for Tool Policy Optimization}
While Supervised Fine-Tuning (SFT) can teach models to mimic tool-use trajectories, Reinforcement Learning (RL) is superior for strategic decision-making in unseen scenarios. The latest SOTA models, such as Llama 3 \citep{dubey2024llama}, heavily rely on advanced RLHF techniques to refine their policy. Building on foundational work demonstrating that RL can enhance reasoning \citep{guo2025deepseek}, recent research has shown that RL teaches not just the "syntax" of tool calling, but the "strategy" \citep{feng2025retool, jin2025search}.

The field is advancing towards more sophisticated, outcome-driven optimization loops. For instance, AlphaCodium \citep{ridnik2024code} employs a test-based, multi-stage iterative process. Similarly, influential frameworks like Tree of Thoughts (ToT) \citep{yao2023tree}, which explores multiple reasoning paths, and Reflexion \citep{shinn2023reflexion}, which leverages self-reflection, highlight the trend towards more deliberate and verifiable reasoning. Our hierarchical reward function aligns with this trend by providing a structured, outcome-driven signal to teach the model a robust planning strategy.

\subsection{Benchmarks for Tool Calling}
The evaluation of LLM tool-calling capabilities relies on increasingly sophisticated benchmarks, which have evolved from early interactive environments like ALFWorld \citep{shridhar2020alfworld} and WebShop \citep{yao2022webshop}. In recent years, benchmarks have grown in scale and realism, with examples like ToolBench \citep{qin2023toolllm} and its more stable successor, StableToolBench \citep{guo2024stabletoolbench}. The focus has also expanded towards greater realism, as seen in the web environment of WebArena \citep{zhou2023webarena}, and more challenging tasks, as in the GAIA benchmark \citep{mialon2023gaia}.

To support this research, new platforms are emerging that focus on specific aspects like evaluating the correctness of API calls in API-Bank \citep{li2023api} and facilitating multi-agent collaboration in open-source frameworks like MetaGPT \citep{hong2023metagpt}. Our work contributes a new benchmark, ComplexTool-Plan, which fills a specific gap in this landscape by targeting the evaluation of an agent's ability to generate complex, non-linear plans.

\section{Methodology}

Our method trains a language model to plan complex multi-tool tasks, as shown in Fig.~\ref{fig:global_framework}. The framework includes: a formal \textbf{Problem Formulation}, our \textbf{ComplexTool-Plan} data generation pipeline, and a \textbf{Model Training} process using both Supervised Fine-Tuning (SFT) and Reinforcement Learning (RL).

\subsection{Problem Formulation}

We formalize multi-tool task planning as a structured prediction task of learning a policy $\pi: Q \times T \rightarrow \mathcal{G}$. Given a natural language query $Q$ and a toolset $T$, the policy generates an execution plan as a Directed Acyclic Graph (DAG), $G=(V, E)$, where vertices $V \subseteq T$ are tools and edges $E$ represent data dependencies. The goal is to train a model $M_\theta$ to find the optimal policy $\pi^*$ that maximizes the expected plan utility $U(G)$:
\begin{equation} \label{eq:objective}
    \theta^* = \arg\max_{\theta} \mathbb{E}_{(Q, T) \sim p(Q,T)} [U(M_{\theta}(Q, T))]
\end{equation}
The utility $U(\cdot)$ is realized through a reward function.

\subsection{ComplexTool-Plan}

To overcome data scarcity, we developed ComplexTool-Plan, a three-stage automated pipeline to generate our training dataset, $D_{train}$.

\begin{itemize}
    \item \textbf{Workflow Generation}: We initiate the process by leveraging a powerful LLM-DeepSeek-V3, to author a large set of workflows. For each workflow, the model is provided with a subset of tools from our predefined library $\mathcal{T}$ and prompted to generate a structurally complex and logically consistent execution plan in the form of a DAG, $G_{orig}$. This LLM-driven approach produces the diverse, high-quality ground-truth solutions for our planning tasks.

    \item \textbf{Query Reverse-Engineering}: Next, we employ a powerful teacher LLM ($M_{teacher}$)-DeepSeek-V3 to reverse-engineer a natural language query $Q$ for each generated workflow $G_{orig}$. The model's objective is to capture the user's intent behind the workflow, effectively transforming the difficult problem of plan generation into a more manageable text-to-text task.

    \item \textbf{Intent Analysis and Re-planning}: The final stage acts as a crucial quality filter. The reverse-engineered query $Q$ might be ambiguous or fail to fully capture the intent of the original plan $G_{orig}$. Therefore, we use the same teacher model $M_{teacher}$, now acting as an expert planner, to re-solve the problem based solely on $Q$. This ensures that the final DAG plan, $G_{final}$, is a faithful and optimal solution derivable from the query, thus guaranteeing the high fidelity of our resulting ($Q$, $G_{final}$) training pairs in $D_{train}$.
\end{itemize}

\subsection{Model Training}

Our training process commences with Supervised Fine-Tuning (SFT) on a suite of Qwen3 models---specifically, the 0.6B, 1.7B, 4B, and 8B variants. This initial SFT phase serves to provide an effective initialization (i.e., a cold start), which is subsequently followed by a GRPO training phase for further policy refinement.

\subsubsection{Base Models and Supervised Fine-Tuning (SFT)}
We first perform Supervised Fine-Tuning (SFT) on our dataset $D_{\text{train}}$ to initialize the model. The objective is to learn to generate the ground-truth plan $G_{gt}$ by minimizing the Negative Log-Likelihood (NLL) loss:
\begin{equation}
    \mathcal{L}_{\text{SFT}}(\theta) = -\mathbb{E}_{(Q, G_{gt}) \sim D_{\text{train}}} [\log P(G_{gt} | Q, T; \theta)]
\end{equation}

\subsubsection{RL Training Set Curation}
\label{sec:rl_training_set_curation}
To ensure a stable and efficient RL phase, we curate the training data to focus on problems at the frontier of the model's current capabilities \citep{chen2024self}. We use the SFT-trained model as a filter: tasks that the model already consistently solves (offering no learning signal) or consistently fails (intractably hard) are excluded from the RL training set. This curation of high-variance instances, where the outcome is uncertain, concentrates the training on the most informative samples, preventing policy degradation and fostering robust optimization.

\subsubsection{Reinforcement Learning with Hierarchical Rewards}

To move beyond SFT's limitations, we use Reinforcement Learning (RL). This, however, requires a nuanced reward signal to properly evaluate complex DAGs, distinguishing structural from strategic errors. We address this by introducing a Hierarchical Reward Function $R(y)$. To optimize this complex, multi-dimensional reward, we utilize the Group Relative Policy Optimization (GRPO) algorithm \citep{shao2024deepseekmath}, which is specifically designed for such scenarios and ensures robust policy improvement.

Our reward function, $R(y)$, where $y$ represents the generated plan, evaluates plan quality through a fail-fast hierarchical process. It prioritizes structural correctness, with critical errors incurring large negative penalties and terminating evaluation.

\begin{itemize}
    \item \textbf{Level 1 and 2: Structural and Semantic Penalties.} We first apply penalties for critical errors that render a plan invalid. This includes ill-formed syntax (e.g., non-JSON) or the presence of a \textbf{cycle}, both resulting in a large penalty (-10.0). A less severe penalty (-2.0) is given for a lack of \textbf{connectivity} (e.g., isolated nodes), which indicates a semantic flaw.
    
    \item \textbf{Level 3: Planning Fidelity Rewards.} If a plan passes all structural checks, it receives a positive reward based on its quality. This reward consists of two parts: 
    (\textit{i}) an \textbf{Edge F1 Score Reward} ($R_{F1}$), calculated as 5 times the edge-level F1 score against the ground truth to reward partial structural correctness, and 
    (\textit{ii}) a large \textbf{Perfect Match Bonus} ($R_{bonus}$) of +5.0 if the plan is identical to the ground truth, incentivizing perfect solutions.
\end{itemize}

The final reward, $R(y)$, is computed based on these components. If any penalty is applied, the reward is the value of that penalty, checked in a specific order of precedence. Otherwise, it is the sum of the fidelity rewards:
\begin{equation}
    R(y) = 
    \begin{cases}
        P_{\mathrm{syntax}}(y) & \text{if invalid syntax} \\
        P_{\mathrm{cycle}}(y) & \text{else if has cycle} \\
        P_{\mathrm{connectivity}}(y) & \text{else if disconnected} \\
        R_{F1}(y) + R_{\mathrm{bonus}}(y) & \text{otherwise}
    \end{cases}
\end{equation}
The domain of this reward function, $[-10.0, 10.0]$, provides a rich, multi-faceted learning signal that guides the model not only on \textit{whether} its plan is correct, but also on the \textit{nature} and \textit{severity} of its errors.

\begin{figure}[htbp]
\centering
\includegraphics[width=\columnwidth]{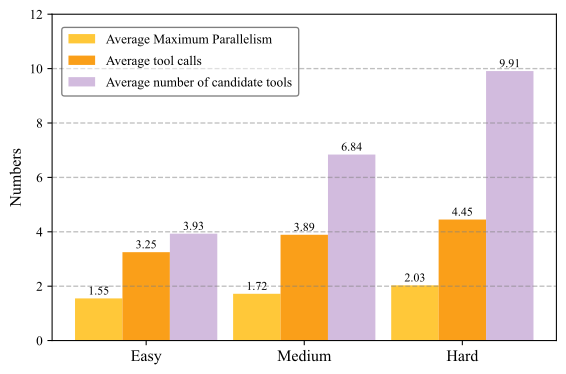} 
\caption{This chart shows our three task difficulties: Easy, Medium, and Hard. Harder tasks have more available tools to choose from (blue) and also require more tools to be used (orange).}
\label{fig:dataset_analysis}
\end{figure}

\section{Experimental Setup}
\subsection{Dataset}
We conduct a comprehensive evaluation of our framework from two primary perspectives: (1) the quality of the generated plans, and (2) the end-to-end task execution success rate. For the former, we use our newly constructed ComplexTool-Plan benchmark. For the latter, we evaluate our system on the established StableToolBench benchmark.

\paragraph{ComplexTool-Plan Benchmark}
We evaluate our Planner on ComplexTool-Plan, our benchmark constructed from a comprehensive library of 4,535 tool APIs sourced from ModelScope~\cite{li2023modelscope}. The benchmark features three difficulty levels (Easy, Medium, and Hard) and provides a curated SFT dataset of 3,000 instances. For RL training, the dataset is filtered to comprise 787 instances, which are subsequently divided into training and test sets using an 8:2 ratio. Further details are available in Fig.~\ref{fig:dataset_analysis}.

\paragraph{End-to-End Execution Benchmark}
To evaluate the practical effectiveness of our framework, we use StableToolBench \citep{guo2024stabletoolbench} for end-to-end evaluation. StableToolBench is an enhanced version of the test set from ToolBench \citep{qin2023toolllm}, specifically designed to mitigate execution instability via a caching system and API simulators. It categorizes test cases across two axes: tool generalization and scenario complexity.
\begin{itemize}
    \item \textbf{Tool Generalization}: \textbf{Inst.} (unseen instructions for seen tools), \textbf{Tool} (unseen tools in seen categories), and \textbf{Cat.} (unseen tool categories).
    \item \textbf{Scenario Complexity}: \textbf{I1} (single-tool), \textbf{I2} (multi-tool in-category), and \textbf{I3} (multi-tool in-collection).
\end{itemize}
Task difficulty escalates from I1-Inst. to I3-Cat., providing a comprehensive testbed for agentic capabilities.

In our end-to-end evaluation on StableToolBench, we pair our Planner with GPT-4o as the executor.

\subsection{Baselines}
To ensure a comprehensive evaluation, we benchmark our framework against two categories of state-of-the-art baselines:

\begin{itemize}
    \item \textbf{Proprietary Planners:} For direct planning evaluation on our ComplexTool-Plan benchmark, we compare against powerful proprietary models prompted to generate DAG plans directly. These include GPT-4o, Claude-3.7, DeepSeek-V3, and Ernie-X1, which serve as a practical upper bound for the task.

    \item \textbf{Open-Source Frameworks:} For end-to-end evaluation on StableToolBench \cite{guo2024stabletoolbench}, we compare against established methods applied to the ToolLLaMA model \cite{qin2023toolllm}. This includes the reactive framework ReAct, the planning-based method DFSDT, and a highly relevant concurrent work, LLMCompiler \cite{kim2024llm}.
\end{itemize}

\subsection{Evaluation Metrics}

\subsubsection{Planning Quality Metrics}
\label{ssubsec:planning_metrics}
To comprehensively evaluate planning quality, we assess three key aspects: the correctness of tool selection (node-level), the accuracy of inferred dependencies (edge-level), and the holistic accuracy of the entire plan. Our evaluation is conducted on a test set comprising 500 samples from the Easy split and 500 from the Hard split. For each generated Directed Acyclic Graph (DAG), we compute the following metrics against the ground-truth plan, with all final scores macro-averaged.

Let $S_{pred}$ be the set of predicted items (nodes or edges) and $S_{true}$ be the set of ground-truth items.

\begin{itemize}
    \item \textbf{Node-level Performance (P/R/F1):} Measures the accuracy of tool selection. We compute Precision, Recall, and F1-score between the set of predicted nodes ($N_{pred}$) and ground-truth nodes ($N_{true}$).
    \begin{itemize}
        \item \textbf{Precision (P)}: Measures fidelity by calculating the fraction of predicted tools that are correct.
        $$ \text{P} = \frac{|N_{pred} \cap N_{true}|}{|N_{pred}|} $$
        \item \textbf{Recall (R)}: Measures completeness by calculating the fraction of required tools that were identified.
        $$ \text{R} = \frac{|N_{pred} \cap N_{true}|}{|N_{true}|} $$
        \item \textbf{F1-Score}: The harmonic mean of Precision and Recall.
        $$ \text{F1} = 2 \cdot \frac{\text{P} \cdot \text{R}}{\text{P} + \text{R}} $$
    \end{itemize}

    \item \textbf{Edge-level Performance (P/R/F1):} Measures the structural correctness of the plan. We apply the same P/R/F1 metrics to compare the set of predicted edges ($E_{pred}$) against the ground-truth edges ($E_{true}$). This assesses the model's ability to infer the correct execution flow.

    \item \textbf{DAG Exact Match (EM):} Our most stringent metric, measuring the holistic correctness of the plan. A plan receives a score of 1 if and only if both its node set and edge set perfectly match the ground truth ($N_{pred} = N_{true}$ and $E_{pred} = E_{true}$), and 0 otherwise.
\end{itemize}

\subsubsection{End-to-End Evaluation Metrics}
\label{ssubsec:e2e_metrics}

To evaluate the end-to-end task-solving capabilities of our framework, we adopt the two primary metrics established by the StableToolBench benchmark. The first, Solvable Pass Rate (SoPR), serves as the primary measure of absolute capability and is defined as the percentage of tasks that an agent successfully completes according to the benchmark's ground-truth criteria. The second, Solvable Win Rate (SoWR), is a comparative metric that evaluates performance relative to a strong baseline. Specifically, it measures the win rate of our agent against the benchmark's provided GPT-3.5 (ReAct) baseline, offering a clear signal of relative improvement. To ensure statistical robustness and mitigate performance fluctuations, all reported scores are the average of three independent runs for each experimental setting.

\begin{table*}[t] 
    \centering

    \begin{threeparttable}
        \begin{tabular}{lccccccc}
            \toprule 
            \multirow{2}{*}{\textbf{Method}} & \multicolumn{3}{c}{\textbf{Node-level Performance}} & \multicolumn{3}{c}{\textbf{Edge-level Performance}} & {\multirow{2}{*}{\begin{tabular}[c]{@{}c@{}}\textbf{DAG Exact}\\\textbf{Match Acc.}\end{tabular}}} \\
            \cmidrule(lr){2-4} \cmidrule(lr){5-7} 
            & {\textbf{Precision}} & {\textbf{Recall}} & {\textbf{F1-Score}} & {\textbf{Precision}} & {\textbf{Recall}} & {\textbf{F1-Score}} & \\
            \midrule 
            GPT-4o          & 0.916 & 0.943   & 0.929 & 0.765   & 0.793 & 0.779   & 0.635   \\
            Deepseek-V3        & 0.777 & 0.763   & 0.770 & 0.641   & 0.645 & 0.643   & 0.511   \\
            Claude-3.7        & 0.939 & 0.959   & 0.949 & 0.801   & 0.829 & 0.815   & 0.644   \\
            Ernie-X1        & 0.925 & 0.973   & 0.948 & 0.716   & 0.744 & 0.730   & 0.562   \\
            \midrule 
            Qwen3-0.6B (SFT)     & 0.951 & 0.986   & 0.968 & 0.826   & 0.872 & 0.848   & 0.671   \\
            Qwen3-1.7B (SFT)     & 0.960 & 0.987   & 0.973 & 0.856   & 0.890 & 0.873   & 0.723   \\
            Qwen3-1.7B (SFT+RL)  & 0.974 & 0.985   & 0.979 & 0.874   & 0.885 & 0.879   & 0.756   \\
            Qwen3-4B (SFT)       & 0.972 & 0.986   & 0.979 & 0.882   & 0.895 & 0.888   & 0.768   \\
            Qwen3-4B (SFT+RL)    & 0.978 & 0.988   & 0.983 & 0.887   & 0.898 & 0.892   & 0.778   \\
            Qwen3-8B (SFT)       & 0.974 & \textbf{0.992}   & 0.983 & 0.889   & \textbf{0.909} & 0.899   & 0.781   \\ 
            Qwen3-8B (SFT+RL)    & \textbf{0.982} & 0.987   & \textbf{0.984} & \textbf{0.905}   & 0.907 & \textbf{0.906}   & \textbf{0.803}   \\ 
            \bottomrule 
        \end{tabular}
    \end{threeparttable}
    \caption{Results of our Planner models and proprietary baselines on the Easy evaluation set. Metrics include node-level, edge-level, and DAG Exact Match accuracy. Our fine-tuned models show superior performance compared to the baselines, with performance scaling positively with model size and the addition of RL. The Qwen3-0.6B(SFT+RL) variant is excluded due to training instability.}
    \label{tab:easy}

    \begin{threeparttable}
        \begin{tabular}{lccccccc}
            \toprule 
            \multirow{2}{*}{\textbf{Method}} & \multicolumn{3}{c}{\textbf{Node-level Performance}} & \multicolumn{3}{c}{\textbf{Edge-level Performance}} & {\multirow{2}{*}{\begin{tabular}[c]{@{}c@{}}\textbf{DAG Exact}\\\textbf{Match Acc.}\end{tabular}}} \\
            \cmidrule(lr){2-4} \cmidrule(lr){5-7} 
            & {\textbf{Precision}} & {\textbf{Recall}} & {\textbf{F1-Score}} & {\textbf{Precision}} & {\textbf{Recall}} & {\textbf{F1-Score}} & \\
            \midrule 
            GPT-4o          & 0.850 & 0.863   & 0.856 & 0.447   & 0.482 & 0.464   & 0.098   \\
            Deepseek-V3        & 0.842 & 0.787   & 0.814 & 0.414   & 0.415 & 0.414   & 0.082   \\
            Claude-3.7        & 0.884 & 0.910   & 0.897 & 0.476   & 0.508 & 0.491   & 0.106   \\
            Ernie-X1        & 0.862 & 0.844   & 0.853 & 0.400   & 0.411 & 0.405   & 0.052   \\
            \midrule 
            Qwen3-0.6B (SFT)     & 0.891 & 0.866   & 0.878 & 0.561   & 0.541 & 0.551   & 0.160   \\
            Qwen3-1.7B (SFT)     & 0.899 & 0.880   & 0.889 & 0.599   & 0.591 & 0.595   & 0.213   \\
            Qwen3-1.7B (SFT+RL)  & 0.914 & 0.861   & 0.887 & 0.616   & 0.579 & 0.597   & 0.218   \\
            Qwen3-4B (SFT)       & 0.918 & 0.894   & 0.906 & 0.632   & 0.623 & 0.627   & 0.241   \\
            Qwen3-4B (SFT+RL)    & 0.918 & 0.876   & 0.896 & 0.631   & 0.611 & 0.621   & 0.259   \\
            Qwen3-8B (SFT)       & 0.925 & \textbf{0.896}   & \textbf{0.910} & 0.661   & \textbf{0.654} & 0.657   & 0.295   \\ 
            Qwen3-8B (SFT+RL)    & \textbf{0.937} & 0.873   & 0.904 & \textbf{0.678}   & 0.641 & \textbf{0.659}   & \textbf{0.319}   \\ 
            \bottomrule 
        \end{tabular}
    \end{threeparttable}
    \caption{Results of our Planner models and proprietary baselines on the Hard evaluation set. Our fine-tuned models show superior performance compared to the baselines, with performance scaling positively with model size and the addition of RL. The Qwen3-0.6B(SFT+RL) variant is excluded due to training instability.}
    \label{tab:hard}

\end{table*}

\begin{table*}[t]
\centering

\setlength{\tabcolsep}{2.1pt}

\begin{threeparttable}
\begin{tabular}{l|cc|cc|cc|cc|cc|cc|cc}
\hline
\multirow{2}{*}{\textbf{Method}} & 
\multicolumn{2}{c}{\textbf{I1-Inst.}} & 
\multicolumn{2}{c}{\textbf{I1-Tool}} & 
\multicolumn{2}{c}{\textbf{I1-Cat.}} & 
\multicolumn{2}{c}{\textbf{I2-Inst.}} & 
\multicolumn{2}{c}{\textbf{I2-Cat.}} &  
\multicolumn{2}{c}{\textbf{I3-Inst.}} & 
\multicolumn{2}{c}{\textbf{Average}} \\

\cline{2-15}
& SoPR & SoWR & SoPR & SoWR & SoPR & SoWR & SoPR & SoWR & SoPR & SoWR & SoPR & SoWR & SoPR & SoWR \\
\hline
\multicolumn{15}{c}{\textbf{GPT-series}} \\
\hline
GPT-3.5 (ReAct)         & 53.0 & --   & 53.0 & --   & 51.2 & --   & 37.6 & --   & 43.9 & --   & 48.6 & --   & 47.9 & --   \\
GPT-3.5 (DFSDT)         & 63.8 & 58.9   & 73.9 & 65.8   & 65.8 & 60.1   & 57.1 & 72.6   & 69.8 & 68.5   & 69.9 & 67.2   & 66.7 & 65.5   \\
GPT-4 (ReAct)           & 54.4 & 53.4 & 44.1 & 60.1 & 48.8 & 52.9 & 50.6 & 69.8 & 48.9 & 62.1 & 42.6 & 54.1   & 48.2 & 58.7   \\
GPT-4 (DFSDT)           & 69.0 & 57.1 & 69.6 & 66.5 & 68.1 & 61.4 & 70.8 & 73.6 & 68.0 & 62.9 & 76.0 & 63.9   & 70.3 & 64.2   \\
\hline
\multicolumn{15}{c}{\textbf{Open-source}} \\
\hline
ToolLLaMA (ReAct)       & 42.7 & 36.2 & 35.4 & 36.1 & 38.6 & 34.6 & 39.9 & 49.1 & 40.9 & 38.7 & 29.8 & 41.0   & 37.9 & 39.3   \\
ToolLLaMA$\dagger$ (ReAct) & 26.7 & 22.1 & 25.0 & 27.2 & 31.7 & 29.4 & 23.1 & 32.1 & 24.5 & 28.2 & 20.5 & 24.6   & 25.3 & 27.3   \\
ToolLLaMA (DFSDT)       & 56.6 & 39.9 & 55.5  & 46.8 & 56.5  & 41.8 & 49.7 & 53.8 & 53.4 & 49.2 & 53.6   & 50.8 & 54.2  & 47.1  \\
ToolLLaMA$\dagger$ (DFSDT) & 41.8 & 35.6 & 39.9 & 37.3 & 44.9 & 39.9 & 36.0 & 47.2 & 39.1 & 39.5 & 33.3 & 26.2   & 39.2 & 37.6   \\
LLMCompiler      & 39.2 & 35.6 & 35.1 & 36.0 & 39.8 & 35.3 & 37.5 & 45.6 & 38.4 & 38.1 & 27.0 & 36.5 & 36.2   & 37.9    \\
Qwen3-1.7B (RL)   & 61.4 & 52.1 & 51.7 & 42.5 & 58.1 & 49.3 & 48.7 & 45.8 & 46.8 & 48.2 & 44.7 & 40.5   & 51.9 & 46.4   \\
Qwen3-4B (RL)    & 62.1 & 54.6 & 55.4 & 47.1 & \textbf{61.4} & \textbf{53.5} & 59.2 & 55.7 & 48.7 & 51.3 & 50.8 & 48.6   & 56.3 & 51.8   \\
Qwen3-8B (RL)    & \textbf{67.0} & \textbf{57.9} & \textbf{56.6} & \textbf{49.8} & 59.3 & 51.7 & \textbf{60.1} & \textbf{58.2} & \textbf{54.7} & \textbf{56.4} & \textbf{61.0} & \textbf{55.9}   & \textbf{59.8} & \textbf{55.0}   \\
\hline
\end{tabular}

\end{threeparttable}

\caption{End-to-end task-solving results on the StableToolBench benchmark. The table compares our Qwen3 models with GPT-series and other open-source methods across various task complexities. Our Qwen3 series, particularly the 8B model, demonstrates the best performance among all open-source baselines in both absolute success (SoPR) and win rate against GPT-3.5(ReAct) (SoWR).}
\label{tab:performance_stabletoolbench}
\end{table*}

\section{Results and Discussion}

\subsection{RQ1: How does Reinforcement Learning enhance performance over Supervised Fine-Tuning?}
Our results demonstrate that the RL stage systematically improves holistic plan accuracy over SFT alone, as measured by our most direct metric: DAG Exact Match Accuracy. As shown in Tab.~\ref{tab:easy}, on the Easy set, our Qwen3-8B model's accuracy increases from \textbf{0.781} (SFT) to \textbf{0.803} (SFT+RL). This trend is consistent across all model sizes and becomes even more critical as task complexity increases. Notably, the training instability of our smallest model (Qwen3-0.6B) suggests this trend has a lower bound: sufficient model capacity is crucial to prevent reward hacking, where a model learns a degenerate, low-effort policy simply to avoid penalties rather than solving the task.

On the challenging \textbf{Hard set} (Tab.~\ref{tab:hard}), the impact of RL is more pronounced. The Qwen3-8B model's DAG Exact Match accuracy jumps from \textbf{0.295} (SFT) to \textbf{0.319} (SFT+RL), an \textbf{8.1\%} relative improvement. This shows RL is not merely polishing plans but is essential for correcting subtle structural errors that emerge in complex scenarios, a conclusion supported by consistent gains in Edge-level F1-Scores. Essentially, while SFT effectively teaches the model to select the right tools (nodes), RL is crucial for orchestrating them correctly (edges). By optimizing for structural integrity, RL moves the model beyond plausible mimicry to generating functionally correct plans.

\subsection{RQ2: How does model scaling affect performance robustness as task complexity increases?}

Our experiments reveal a clear, positive relationship between model size and planning capability, with larger models demonstrating significantly greater robustness against increasing task complexity.

First, our framework shows strong scalability. This is most evident on the challenging Hard set, where even top-tier models like GPT-4o falter. On this set, the DAG Exact Match Accuracy of our SFT+RL models rises monotonically with scale: from \textbf{0.218} (1.7B) to \textbf{0.259} (4B) and finally \textbf{0.319} (8B). This stark performance gap over general-purpose models validates our specialized training approach. 

Second, and more importantly, model scaling directly enhances robustness. While all models degrade on harder tasks, larger models exhibit a more graceful decline. For instance, when moving from the Easy to the Hard set, the accuracy of our Qwen3-1.7B model plummets by \textbf{71.2\%} (from 0.756 to 0.218). In contrast, our largest Qwen3-8B model sees a more contained \textbf{60.3\%} drop (from 0.803 to 0.319). This widening performance gap is crucial, demonstrating that scaling up our specialized Planner not only boosts absolute performance but also enhances its \textbf{resilience} against complex, real-world challenges.

\subsection{RQ3: What is the end-to-end effectiveness and efficiency of our framework?}
\label{sec:rq3}

Our framework demonstrates highly competitive end-to-end performance. As shown in Tab.~\ref{tab:performance_stabletoolbench}, our Qwen3-8B (RL) model achieves an average SoPR of \textbf{59.8\%}, substantially outperforming the widely-used reactive approach of GPT-4 (ReAct) at 48.2\%. This highlights that a superior planning strategy can be more impactful than raw model capability, unlocking the potential of smaller models.

While iterative paradigms like DTA-Llama~\cite{zhu2025divide} report a higher SoPR, their success is largely driven by a multi-turn framework that allows for in-process error correction and adaptation. In contrast, our non-iterative, plan-then-execute paradigm has only one chance to generate a correct global plan upfront. This architectural choice prioritizes predictability and planning quality over the adaptive resilience of iterative feedback loops.

Crucially, this design choice leads to state-of-the-art efficiency. As detailed in Tab.~\ref{tab:inference_steps}, our method requires an average of just \textbf{2.29 inference steps} to complete a task, significantly fewer than all other methods, including iterative parallel frameworks like DTA-Llama (2.48 steps). This demonstrates that our global planning approach is not only effective but also architecturally more efficient, solving complex tasks with fewer high-level decision rounds.

\begin{table}[t]
\centering
\small 
\setlength{\tabcolsep}{2.8pt} 

\begin{threeparttable}
\begin{tabular}{l|cccccc}
\hline
\textbf{Method} & \textbf{I1-I.} & \textbf{I1-T} & \textbf{I1-C.} & \textbf{I2-I.} & \textbf{I2-C.} & \textbf{I3-I.} \\
\hline
\multicolumn{7}{c}{\textbf{GPT-series}} \\
\hline
GPT-3.5 (ReAct)         & 4.28 & 4.75 & 4.48 & 5.16 & 5.05 & 5.31 \\
GPT-3.5 (DFSDT)         & 11.60 & 13.36 & 11.77 & 16.60 & 14.06 & 12.54 \\
GPT-3.5 (Parallel)      & 25.33 & 28.06 & 26.12 & 31.79 & 31.04 & 38.10 \\
GPT-4 (ReAct)           & 3.27 & 3.64 & 3.87 & 4.04 & 4.19 & 4.23 \\
GPT-4 (DFSDT)           & 5.90 & 8.09 & 6.67 & 9.97 & 18.13 & 14.05 \\
GPT-4 (Parallel)           & 4.66 & 9.18 & 12.90 & 3.63 & 5.98 & 10.38 \\
\hline
\multicolumn{7}{c}{\textbf{Open-source}} \\
\hline
ToolLLaMA (ReAct)       & 3.42 & 3.47 & 3.50 & 3.67 & 3.63 & 3.64 \\
ToolLLaMA (DFSDT)       & 8.09 & 8.51 & 8.10 & 10.20 & 9.93 & 9.23 \\
LLMCompiler             & 5.48 & 5.56 & 6.07 & 5.36 & 5.68 & 5.62 \\
Qwen2.5 (Parallel)      & 9.07 & 9.47 & 12.01 & 14.58 & 14.56 & 12.38 \\
DTA-Llama             & 2.41 & 2.41 & 2.51 & 2.32 & 2.34 & 2.48 \\
Qwen3-8B (RL)           & \textbf{2.26} & \textbf{2.33} & \textbf{2.29} & \textbf{2.16} & \textbf{2.29} & \textbf{2.41} \\
\hline
\end{tabular}
\end{threeparttable}

\caption{Inference steps for different methods. }
\label{tab:inference_steps}
\end{table}

\section{Conclusion}

In this work, we introduced a Planner-centric framework that decouples planning from execution to address the local optimization traps of reactive tool-use agents. Our core contribution is a Planner model trained via a two-stage SFT-GRPO strategy on our new \textbf{ComplexTool-Plan} benchmark. It generates a global, parallelizable DAG plan in a single forward pass.

Our experiments demonstrate that this paradigm is highly effective. The Planner itself produces higher-quality plans than strong proprietary models. When integrated with an executor, our framework achieves state-of-the-art results for open-source models on the challenging \textbf{StableToolBench} benchmark, while requiring the fewest inference steps.

Crucially, our findings show that a sophisticated planning strategy can be more impactful than raw model capability, offering a more scalable and predictable path toward capable autonomous agents.

\bibliography{aaai2026}

\end{document}